\documentclass[11pt]{article}

\usepackage[final]{acl}

\usepackage{times}
\usepackage{latexsym}

\usepackage[T1]{fontenc}

\usepackage[utf8]{inputenc}

\usepackage{microtype}

\usepackage{inconsolata}

\usepackage{graphicx}

\usepackage{booktabs} 
\usepackage{multirow} 
\usepackage{makecell} 
\usepackage{amsmath} 
\usepackage{fancyvrb}
\usepackage{verbatim}
\title{AgentArch: A Benchmark for Evaluating Agent Architectures in Enterprise
Workflows}

\author{
 \textbf{Tara Bogavelli},
 \textbf{Hari Subramani},
 \textbf{Roshnee Sharma}
\\
 ServiceNow
\\
 \small{
   \textbf{Correspondence:} \href{tara.bogavelli@servicenow.com}{tara.bogavelli@servicenow.com}
 }\\
 \small{
 \href{https://github.com/ServiceNow/AgentArch}{https://github.com/ServiceNow/AgentArch}
 }
}

\begin{document}
\maketitle
\begin{abstract}
While individual components of agentic architectures have been studied in isolation, we lack empirical evidence on how design dimensions interact in realistic multi-agent systems. This study aims to address these gaps by providing a comprehensive enterprise-specific benchmark. We introduce \textbf{AgentArch}---a benchmark that evaluates 18 agentic configurations across state-of-the-art large language models. The study spans four dimensions: orchestration strategy, agent style (ReAct versus function calling), memory architecture, and thinking tool integration. Results reveal strong model-specific architectural preferences that challenge one-size-fits-all assumptions in agentic AI. They also expose notable performance gaps on enterprise tasks: the best models reach only 35.3\% success on the complex task and 70.8\% on the simpler task. We aim to inform architecting choices for enterprise agents by grounding model selection and component design in systematic evidence.

\end{abstract}

\section{Introduction}

Large language models (LLMs) have evolved beyond simple text generation to function as autonomous agents capable of iterative decision-making and complex task completion \cite{taubench:25}. This transformation has opened new possibilities for enterprise automation, where agentic AI systems can interact with existing workflows, databases, and business processes to accomplish multi-step objectives with minimal human intervention.

Despite growing interest in agentic systems, practitioners face two critical gaps: existing research predominantly evaluates individual components in isolation—orchestration strategies, prompting approaches, memory management, etc—without examining how these architectural choices interact with each other in real-world scenarios. Additionally, while existing benchmarks span diverse domains from gaming to research tasks, few focus specifically on enterprise workflows. Enterprise use cases are uniquely constrained by the need for reliability and seamless integration with existing business systems. The fragmented approach of prior benchmarks leaves enterprise developers without empirical guidance for selecting optimal combinations of architectural components or choosing appropriate models for their specific setups.

This paper presents a comprehensive evaluation of 18 distinct agentic architectures across state-of-the-art LLMs, systematically examining the interactions between four key architectural dimensions: orchestration strategy (single agent vs. multi-agent), agent style (function calling vs. ReAct), memory management (complete vs. summary), and thinking tool integration \cite{react:23}. We test each architectural combination on two enterprise use cases of varying complexity, providing practical insights for agentic system designers.

Our findings reveal significant model-specific architectural preferences, challenging assumptions about universally optimal designs. We find that even between use cases, models perform best on different architectures. While larger models display more robust performance across architectures, we find that on simpler tasks, smaller models can match performance under their best performing architecture. We also observe significant weaknesses across models when ReAct prompting is used in multi-agent systems. Our results provide empirically-grounded guidance for practitioners building enterprise agentic systems, enabling more informed decisions about architectural components and model selection based on specific use case requirements.

To summarize, our work makes the following contributions:

\begin{itemize}
    \item We introduce \textbf{AgentArch}, the first benchmark systematically evaluating 18 agentic architectures across six large language models on realistic enterprise workflows.
    \item We jointly analyze the effects of orchestration, prompting style, memory design, and thinking tool integration, providing quantitative insight into how these factors interact in practice.
    \item Our experiments reveal model-specific architectural preferences, showing that optimal configurations vary by model and task complexity, rather than following a single best-performing design.
    \item  We offer evidence-based recommendations for selecting models and architectural configurations suited to enterprise reliability and integration constraints.
\end{itemize}

Overall, \textbf{AgentArch} contributes a systematic, empirically grounded step toward understanding how architectural design choices influence agentic system behavior, supporting more informed and reproducible development practices.

\begin{table*}[t]
\small
\centering
\begin{tabular}{p{2.5cm}p{5cm}p{7cm}}
\toprule
 & \textbf{Requesting Time Off (TO)} & \textbf{Customer Request Routing (CR)} \\
\midrule
\textbf{Description} & Simple workflow for PTO eligibility and request processing. Tests basic multi-step reasoning and counting. & Complex workflow mimicking customer service systems. Automates simple requests, escalating complex issues. Tests classification and instruction following. \\
\midrule
\textbf{Tools} & 8 custom tools & 31 custom tools \\
\midrule
\textbf{Agents} & Employee Information Retriever, Balance Eligibility, Leave Approval & Request Validation, Duplicate Case Detection, Case Creation, Informational Queries, Transactional Queries, Intent and Sentiment Extraction, Entity Extraction, Document Verification, Email Generation \\
\midrule
\textbf{Key Challenges} & Date calculations; leave balance verification; policy compliance; multi-step approval workflow  & Appropriate escalation decisions; context preservation; handling ambiguous requests; complex routing logic; instruction following \\
\bottomrule
\end{tabular}
\caption{Detailed comparison of the two enterprise use cases evaluated in our benchmark.}
\label{tab:use-cases}
\end{table*}

\section{Related Work}
Evaluating agentic AI systems has become a central research focus, yet many studies still assess components in isolation rather than integrated end-to-end performance.

\textbf{Tool calling.} Tool use is foundational for agentic systems. Benchmarks such as \cite{agentbench:24, bfcl:25, sealtools:24} document persistent weaknesses in tool selection, argument fidelity, output formatting, latency, and redundant calls. More complex settings further degrade performance: combining tool use with intricate instruction following \cite{taubench:25} or multi-step, nested tool invocations \cite{basu2025nestfulbenchmarkevaluatingllms} increases error rates.

\textbf{Memory and context management.} Limited context windows and growing conversational and tool-response histories create pressure on memory design. Prior work explores tiered memory as an OS abstraction \cite{packer2024memgptllmsoperatingsystems}, dynamic knowledge graphs for retrieval and linking \cite{xu2025amemagenticmemoryllm}, and dual-track (short/long-term) schemes \cite{liu2024llmconversationalagentmemory}. Empirical studies highlight how tool calls inflate context via frequent, verbose outputs \cite{maharana2024evaluatinglongtermconversationalmemory}, motivating mechanisms that preserve relevance while controlling prompt length.

\textbf{Multi-Agent Systems (MAS).} With increasing workflow complexity, MAS introduce specialization and coordination. Benchmarks examine planning structures (hierarchical vs.\ distributed) and execution quality \cite{marble:25}, as well as inter-agent planning/coordination \cite{llmcoord:25,geng2025realmbenchrealworldplanningbenchmark}, identifying planning reliability as a bottleneck. Routing frameworks improve agent/model selection and communication patterns \cite{masrouter:25}; adaptive approaches search multi-agent configurations online \cite{hou2025halohierarchicalautonomouslogicoriented}. Automatic scaling from single agents to MAS can yield gains \cite{yuan2025evoagentautomaticmultiagentgeneration}, and heterogeneous-model MAS leverage complementary strengths \cite{ye2025xmasbuildingmultiagentsystems}.

\textbf{Enterprise evaluations.} Enterprise-focused studies consistently reveal a gap between academic benchmarks and real deployments. Even state-of-the-art models struggle on routine workplace tasks \cite{workbench:24}, with performance degrading in multi-turn and interdependent workflows \cite{crmarena:25,huang2025crmarenaproholisticassessmentllm,workarena:24,workbench:24}. These findings underscore the need for evaluations that reflect realistic business constraints: reliability, strict procedure orderings, and integration with existing systems.

Prior work has advanced understanding of individual components—tool use \cite{taubench:25,bfcl:25,basu2025nestfulbenchmarkevaluatingllms}, memory \cite{locomo:24}, and coordination \cite{geng2025realmbenchrealworldplanningbenchmark,llmcoord:25}—but remains fragmented. Enterprise-oriented benchmarks capture task complexity yet rarely analyze \emph{interactions} among orchestration, prompting style, memory strategy, and reasoning aids. Our benchmark fills this gap by jointly varying these dimensions across two enterprise use cases, providing evidence on model-specific architectural preferences and the relative advantages of multi-agent versus single agent designs.

\begin{table}[t]
\centering
\scriptsize
\setlength{\tabcolsep}{5pt}
\begin{tabular}{l@{\hspace{10pt}}rrr@{\hspace{10pt}}rrr}
\toprule
\textbf{Model} & \multicolumn{3}{c}{\textbf{TO}} & \multicolumn{3}{c}{\textbf{CR}} \\
\cmidrule(lr){2-4} \cmidrule(lr){5-7}
 & Mean & SD & CV &Mean & SD & CV \\
\midrule
GPT-4.1        & 48.2 & 13.0 & 27.0 & 16.1 & 5.5 & 34.4 \\
GPT-4o         & 31.4 & 18.7 & 59.7 & 1.8 & 1.5 & 83.9 \\
GPT-4.1-mini   & 38.8 & 22.2 & 57.2 & 1.3 & 1.5 & 110.7 \\
o3-mini        & 15.5 & 22.3 & 143.7 & 9.7 & 10.1 & 104.0 \\
LLaMA 70B      & 1.1 & 3.1 & 286.8 & 0.0 & 0.0 & 0.0 \\
Sonnet 4       & 49.0 & 15.7 & 32.1 & 15.5 & 12.3 & 79.1 \\
\bottomrule
\end{tabular}
\caption{Model consistency analysis - mean, standard deviation, and coefficient of variance for acceptable pass@1 scores. Performance statistics across all 18 architectural configurations for each use case.}
\label{tab:model-consistency-compact}
\end{table}

\section{Methodology}

\subsection{Experiment Set Up}
\textbf{Use Cases}
We evaluate two enterprise workflows of differing complexity to examine how task complexity affects agentic performance on real-world enterprise automation scenarios. Both of these use cases involve highly-specific directions to complete tasks in a given order, reflecting the rigid nature of many enterprise workflows where steps must take place in a predefined order. For each request in a use case, we define an expected final decision (e.g. leave approved or rejected). We create mock data so that each tool request returns a deterministic result. Unlike existing benchmarks that use simplified, clean responses, we deliberately construct enterprise-realistic data that is complex, lengthy, and messy. Knowledge base articles span thousands of words with extensive documentation reflecting real enterprise systems. Tool responses return complex JSON objects where relevant information is often buried within metadata, error codes, and auxiliary fields. This approach better represents the information processing challenges agents face in production environments where data is rarely clean or perfectly structured.

\textbf{Requesting Time Off (TO)} represents a simple workflow involving PTO eligibility verification and request processing. This use case tests basic multi-step reasoning, counting, and decision-making with clear success criteria, employing 8 custom tools and 3 agents.

\textbf{Customer Request Routing (CR)} represents a complex workflow that mimics intelligent customer service systems, handling simple requests automatically while escalating complex issues to human agents. This tests nuanced classification, context understanding, and appropriate escalation decisions with 31 custom tools and 9 agents.
Each use case includes 60 user utterances designed to cover the full spectrum of real-world requests, including successful scenarios, edge cases, and failure conditions that would occur in actual enterprise deployments. Implementation details are provided in Appendix~\ref{sec:usecases}.

\textbf{Orchestration Strategies}.
We evaluate three orchestration strategies to examine single agent versus multi-agent performance across key dimensions: decision-making with varying option sets, collaboration costs and benefits, and specialist versus generalist agent effectiveness. \textbf{1) Orchestrator-led, Isolated Agent.} An orchestrator assigns tasks to specialist agents and controls all inter-agent communication. When agents need additional information, they request help through the orchestrator, who selects an appropriate agent to respond. This tests whether centralized planning and controlled information flow improve outcomes. \textbf{2) Orchestrator-led, Open Agent Network.} The orchestrator assigns initial tasks, but agents communicate directly with each other when additional information is needed. Agents select collaborators independently using agent descriptions. This examines whether direct agent-to-agent communication is more efficient than orchestrator-mediated interaction. \textbf{3) Single Agent.} A single agent receives the user request and has access to all available tools, with no collaboration mechanisms. This tests agent performance in larger, unfiltered option spaces.

\begin{figure}[htbp]
  \centering
  \small
  \includegraphics[width=0.9\columnwidth]{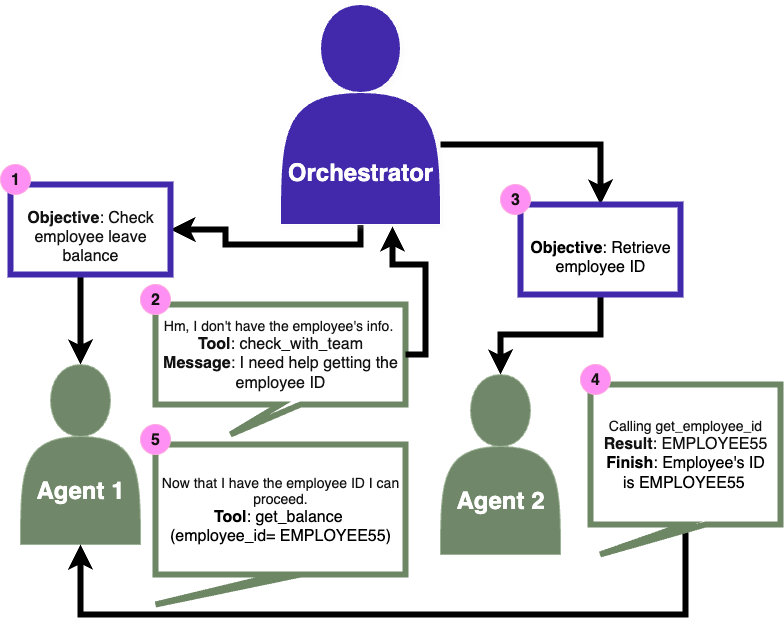}
  \caption{Orchestration, Isolated Agents. Agents ask the orchestrator for help, and the orchestrator selects agent to assist.}
  \label{fig:your-label}
\end{figure}

\textbf{Agents}.
We evaluate two prevalent agent styles to examine how different prompting paradigms influence model performance. With \textbf{function calling}, the model directly selects tools from the available toolkit, leveraging inherent function calling capabilities without explicit intermediate reasoning. Using \textbf{ReAct}, the model outputs its reasoning before selecting actions, following a structured reasoning-action framework \cite{react:23}. We investigate whether the explicit reasoning in ReAct prompting yields substantially different outcomes compared to direct function calling. Implementation details and prompts are provided in the appendix.

\textbf{Memory Management}.
We evaluate two information-sharing approaches to examine the trade-off between context completeness and input context length. With \textbf{complete memory}, agents see all previous tool calls, parameters, and responses from other agents, maximizing context but increasing prompt length. With \textbf{summarized memory}, agents receive only final summaries from previous agents thus reducing context length.This tests whether comprehensive information sharing enhances performance or condensed summaries suffice for effective collaboration.

\textbf{Thinking Tools}.
We evaluate whether providing explicit reasoning tools improves model performance on multi-step analysis tasks. When thinking tools are enabled, models receive access to \textbf{math} for calculations and \textbf{synthesize\_collected\_information} for data organization, particularly relevant for PTO eligibility, where models must calculate date ranges and compare leave balances. We hypothesize that models without built-in reasoning capabilities will benefit from these structured tools. See Appendix A.2 for more details and examples.

\begin{figure}[htbp]
  \centering
  \small
  \includegraphics[width=0.9\columnwidth]{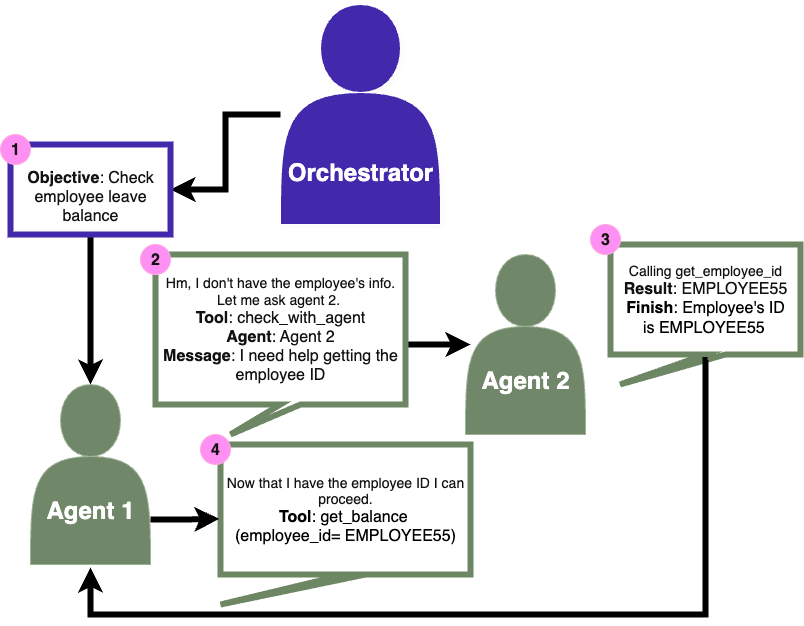}
  \caption{Orchestration, Open Agent Network. Agents ask other agents for help directly.}
  \label{fig:your-label}
\end{figure}

\textbf{Model Settings}. We evaluate our benchmark using six models: OpenAI GPT-4.1, OpenAI-GPT-4o, LLaMA 3.3 70B, OpenAI GPT-4.1-mini, Anthropic Claude Sonnet 4, and OpenAI o3-mini. We evaluated Claude Sonnet 4 with reasoning disabled. For all experiments, we set the temperature to 0 and use no other sampling parameters. We selected this set of models to represent state-of-the-art models of varying sizes, including models with and without in-built reasoning capabilities. For function calling architectures, we use the in-built OpenAI function calling APIs for all models besides Sonnet 4 which uses the Bedrock API.

\begin{figure*}[htbp]
  \centering
  \includegraphics[width=\textwidth]{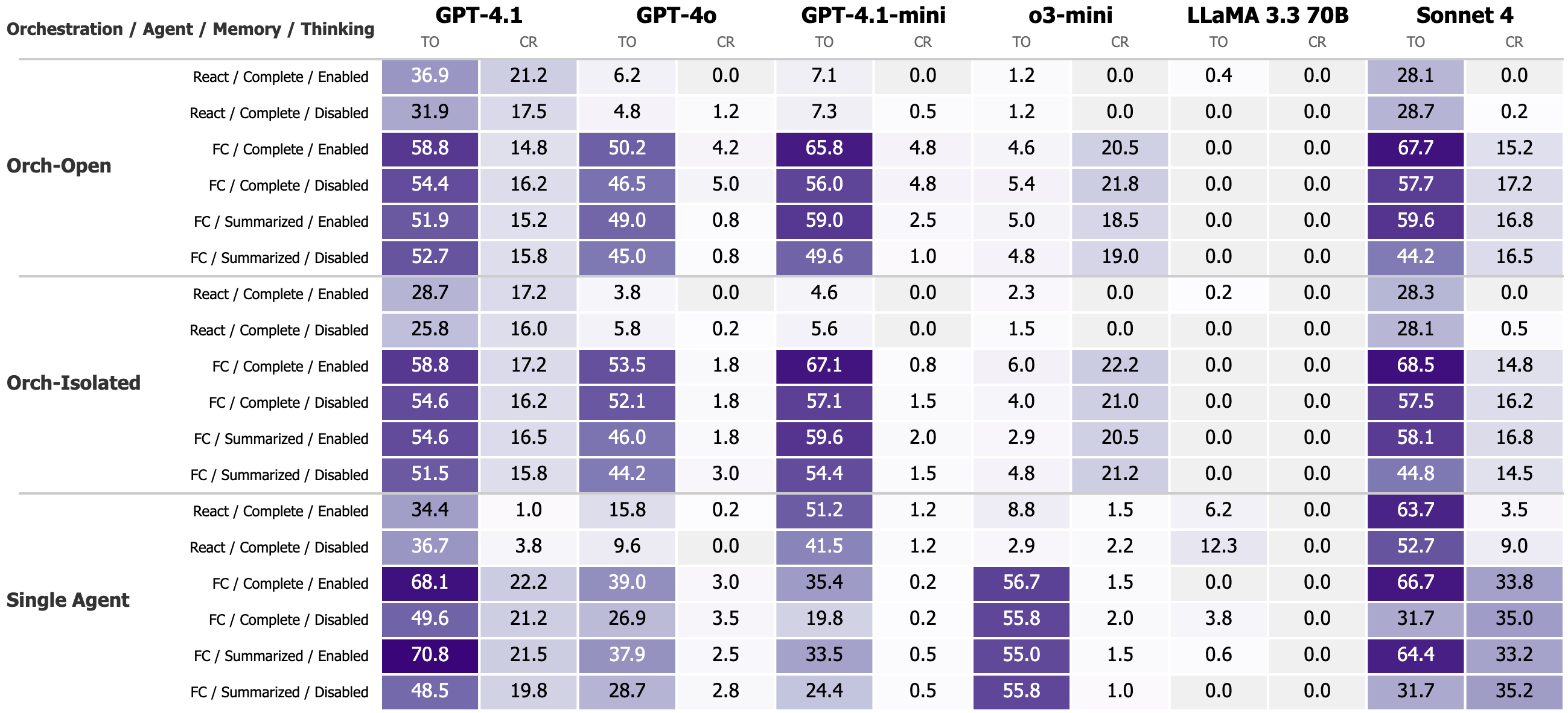}
  \caption{\textbf{Pass@1 Acceptable Rate.} Heatmap cells report the Pass@1 \emph{Acceptable} score (higher/darker is better) for each model–configuration pair on the two enterprise tasks: Time Off (\textbf{TO}) and Customer Routing (\textbf{CR}). Pass@1 is computed over $k=8$ attempts per configuration. Overall, function-calling configurations tend to outperform ReAct, thinking tools often help on the simpler TO task, and no single architecture dominates across models or tasks.}

  \label{fig:your-label}
\end{figure*}

\subsection{Evaluation Metrics}
In enterprise agentic systems, reaching the correct conclusion is necessary but insufficient. Consider a leave request scenario: even if the model correctly determines that leave should be approved, the task fails if the model doesn't invoke the approval API, calls it with incorrect parameters (e.g., wrong employee ID), or omits required notifications. Real-world agentic deployment requires that the agentic system \textbf{(1)} selects the right tools to complete all required steps, \textbf{(2)} invokes them with correct arguments, and \textbf{(3)} reaches the correct final outcome.

As such, our primary evaluation metric is the \textbf{Acceptable Score}, which measures the percentage of records that satisfy all three critical success conditions simultaneously:

$$\textbf{Acceptable} = \frac{|\{r \in R : C(r) \land A(r) \land O(r)\}|}{|R|} \times 100\%$$

where $R$ is the set of all test records, and for each record $r$:
\begin{itemize}
\item $C(r)$ indicates correct tool choice
\item $A(r)$ indicates correct tool arguments ($A(r) = 1.0$)  
\item $O(r)$ indicates correct final decision
\end{itemize}


\textbf{Tool Choice Evaluation.}
We evaluate tool choice under two conditions:

\textit{Lenient Acceptable} ($C_L(r)$): The model executes all required ground truth tools but may perform additional read-only tools. Extraneous write operation tools are penalized.

\textit{Strict Acceptable} ($C_S(r)$): The model must execute exactly the ground truth tools in the correct order with no extraneous tools and zero hallucinations.

\textbf{Tool Arguments Scoring.}
Tool arguments receive percent scoring where $A(r) = 1.0$ only when all arguments are correct (compared to ground truth).

\textbf{Reliability Metrics.}
To assess system reliability, we evaluate each configuration over $k=8$ attempts and compute \textbf{pass@1}—the percent of correct samples over $k$ trials. \cite{deepseekai2025deepseekr1incentivizingreasoningcapability} and \textbf{pass\^{}K}—the probability of \textit{all} $k$ trials succeeding \cite{taubench:25}. The latter metric is particularly important in real-world enterprise domains where reliability and consistency are critical.

We use \textbf{Acceptable pass@1} as our primary metric. We use lenient correct tool choice for the acceptable score as opposed to strict as strict scoring remains extremely challenging for most models. We use \textbf{pass@1} as it brings stability to the lenient acceptable metric over \textit{k} attempts, representing the success rate for a given model and configuration.

\textbf{Additional Metrics}
We track supplementary behavioral metrics, including:
\begin{itemize}
\item \textit{Hallucination rate}: Frequency of selecting non-existent agents or tools
\item \textit{Tool Repetition rate}: Frequency of consecutive identical tool calls with identical arguments
\end{itemize}

\section{Results}
\subsection{Overall Results}

\begin{figure*}[htbp]
  \centering
  \includegraphics[width=\textwidth]{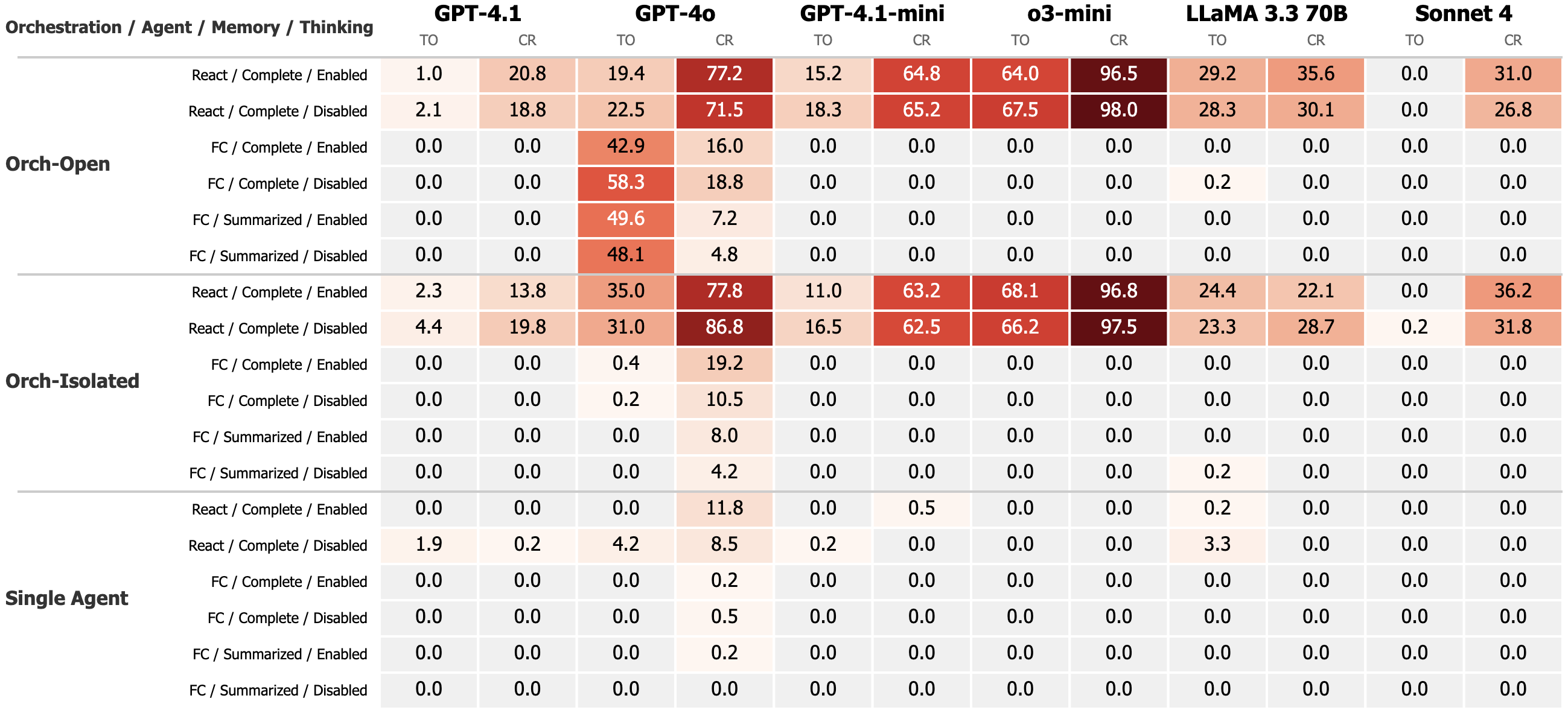}
  \caption{\textbf{Pass@1 Hallucination Rate ($\downarrow$).} Heatmap cells show the percentage of trials (lower is better, darker indicates higher error) in which an agent hallucinated non-existent entities or schema, e.g., selecting a tool or agent not present in the registry or inventing parameters not defined by a tool schema. Rates are computed at Pass@1 over $k=8$ attempts per model–configuration on both tasks (TO, CR). Hallucinations tend to concentrate in ReAct settings, especially in multi-agent orchestration.}
  \label{fig:your-label}
\end{figure*}

Complex enterprise workflows remain challenging for current agentic systems. On the simpler time-off task (TO), peak performance reaches 70.8\% (GPT-4.1), while the more complex customer routing task (CR) peaks at 35.3\% (Sonnet 4). Only GPT-4.1, Sonnet 4, and o3-mini show meaningful capability on CR; most other models score in the single digits. Performance varies markedly across architectures, with no universally best configuration. Models also peak on different configurations across the two tasks, indicating that optimal design depends on task complexity.


With regards to architecture elements, \textbf{thinking tools} consistently improved performance for non-reasoning models on the simpler time off task. GPT-4.1 improved from 48.5\% to 70.8\% with thinking tools enabled in single agent function calling with summarized memory. However, thinking tools provided minimal benefit for o3-mini (55.8\% to 56.7\%) in its best performing category. Thinking tools showed minimal impact on the complex customer routing use case across all models. Compared to other elements, variation in \textbf{memory management} styles and 
\textbf{orchestration strategies} (within multi-agent architectures) had minimal impact on scores.

\textbf{Function calling} generally outperformed \textbf{ReAct} across most models. The notable exception was LLaMA 3.3 70B, which achieved its peak score of 12.2\% using single agent ReAct on the requesting time off task. \textbf{Multi-agent ReAct} consistently underperformed indicating that ReAct may be better suited towards single agent systems. \textbf{Hallucinations} for all models (except for GPT-4o) were found exclusively under ReAct settings suggesting that while models may be robust to hallucinations on the function calling schemas they have been trained on, they still have the propensity to hallucinate in ReAct settings. This is particularly relevant on the more complex use case in multi-agent systems. Sonnet 4, for example, showed hallucination rates of 36-36\% on ReAct in multi-agent systems but 0\% hallucination in every other configuration. See Figure 4 for hallucination rates.

While some models achieved their highest overall scores on single agent architectures, the \textbf{correct final decision} scores for the complex customer routing use case were generally higher for multi-agent architectures. Although Sonnet 4's highest overall scores were all achieved with single agent function calling (all scores above 33\%), its highest correct final decision scores were for function calling with multi-agent systems (84-87\%), while for single agent, it was in the 72-76\% range. GPT-4.1 showed similar results as it achieved its peak overall score in both use cases with single agent function calling; however, on the customer routing task, it achieved correct final decision rates of 97-99\% with multi-agent function calling and only 79-86\% with single agent function calling. GPT-4.1-mini also achieved significantly higher correct final decision rates with multi-agent function calling than with single agent function calling on both use cases. This suggests that while some models are more likely to choose the correct tool or tool arguments on single agent systems, multi-agent systems may be more effective for selecting the correct final decision after all tool calls have been made. See Figure 5 for more details. 

Two key patterns emerged from the \textbf{consistency analysis} shown in Table 2. Sonnet 4 and GPT-4.1 showed the most robust performance with relatively low coefficients of variation (32.1\% and 27.0\%, respectively, for the simpler task), making them the most reliable choices across different agentic configurations. Conversely, o3-mini exhibited extreme sensitivity to architectural choices (CV = 143.7\%).

\textbf{Reliability Analysis}. Pass\^{}k scores across all models and agentic configurations peak at 0.0634, indicating only a 6.34\% chance of executing the workflow correctly in all 8 trials.  In enterprise settings, where reliability and consistency are essential, pass@1 and pass\^{}k results reveal a fundamental gap between the promise of agentic LLMs and their real-world performance. See Appendix A.4 for full scores.

\subsection{Model-Specific Results}

\begin{figure*}[htbp]
  \centering
  \includegraphics[width=\textwidth]{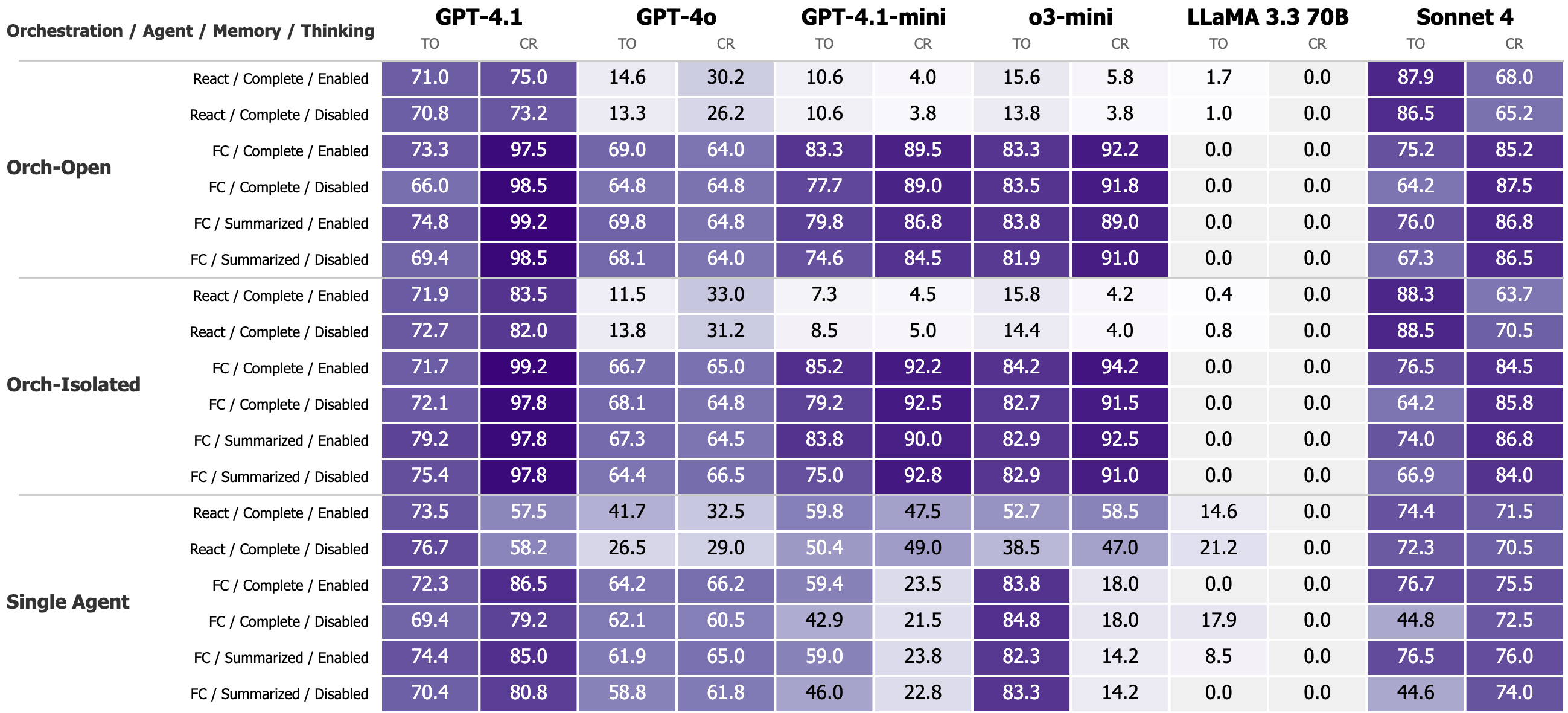}
  \caption{\textbf{Pass@1 Correct Final Decision Rate.} Heatmap cells report the probability (higher/ darker is better) that a model–configuration returns the correct ground-truth outcome (\emph{approve/deny}, \emph{route/escalate}, etc.). Computed at Pass@1 over $k=8$ attempts on both tasks (TO, CR). Together with Fig.~\ref{fig:your-label} (Acceptable), this highlights a trade-off between decision accuracy and end-to-end execution reliability.}

  \label{fig:your-label}
\end{figure*}

Sonnet 4 achieved the highest scores on the complex CR task with a peak score of 35.3\% using single agent function calling. Sonnet 4 performed strongly on the TO task (68.5\% using orchestrator-isolated function calling). Notably, thinking tools had a substantial impact on the simpler task but a minimal effect on the CR task. TO scores improved from 64.4\% to 68.5\% with thinking tools, while CR performance remained relatively flat across configurations (33.3\%-35.3\% range). 

GPT-4.1 achieved the highest overall score on the simpler use case - 70.8\% using single agent function calling with summarized memory and thinking tools. The model showed sensitivity to thinking tools in single agent configurations, improving from 48.5\% to 70.8\% when enabled. However, thinking tools provided minimal benefit in multi-agent setups (54.4\% vs 58.8\%). Memory type preferences varied by orchestration: single agent performance was nearly identical with complete (68.\%1) versus summarized (70.8\%) memory, while orchestrated configurations slightly favored complete memory.

o3-mini demonstrated extreme architectural sensitivity, achieving 56.7\% on time off requests with single agent function calling but dropping to 1.3\% with orchestrated ReAct. Thinking tools showed negligible differences (55.8\% vs 56.7\%). The model achieved its highest CR score of 22.3\% using orchestrator-isolated function calling, substantially outperforming single agent performance (1.5\%-2.0\%). This suggests o3-mini's performance depends heavily on task complexity rather than inherent suitability to single versus multi-agent architecture.

GPT-4o and GPT-4.1-mini peaked with orchestrator-isolated function calling (53.5\% and 67.1\% TO) using full memory and thinking tools. Both gained from thinking tools on simple tasks—GPT-4o rose from 26.9\% to 39.0\% in single-agent setups, though gains faded under orchestration. GPT-4o struggled with ReAct, while GPT-4.1-mini scored higher on ReAct (41.5–51.2\%) than function calling (19.8–35.4\%) in single-agent mode, underscoring how agentic elements interact differently across architectures.

LLaMA 3.3 70B performed poorly overall, scoring 0 on most configurations. Its best result was 12.2\% on the TO task with single-agent ReAct (no thinking tools)—the only model to perform best with ReAct. Scores on customer routing were near zero due to persistent instruction-following issues.

\section{Discussion}
We now use the above findings to make concrete recommendations for designing agentic AI systems in an enterprise.

\textbf{Avoiding Multi-Agent ReAct}.
One of the few universal trends that this benchmark shows is that models do not perform their best on Multi-Agent ReAct. In our setup, we used ReAct both for the orchestrator to select agents and for the agents to select tools. It is possible that using ReAct for just one instead of both of these components might improve outcomes, but further work would be needed to test this.

\textbf{Multi-Agent Systems for Final Decisions}.
While some models performed better overall on single agent systems, we find that multi-agent systems are generally more effective at arriving at the correct final decision. For enterprise use cases where arriving at a final decision is more important than exact tool selection, we recommend the use of multi-agent systems. 

\textbf{Use Case Considerations}.
We observe that models tend to achieve peak performance under different architectures depending on the specific use case, rather than consistently performing best under a single architecture. As such, we caution developers against assuming a single architecture is particularly suited for any model and instead recommend testing against several use cases.

\textbf{Using Thinking Tools}.
Our results indicate that for non-reasoning models, thinking tools can be a useful way for enabling more careful analysis of information collected over the course of an agentic workflow. This is particularly true in contexts where the model must make a calculated decision based on the information it has collected (as is the case with the requesting time off use case). Thinking tools, however, also add latency, so we recommend developers consider carefully if their use cases require the extra reasoning abilities that these tools provide. 

\textbf{Model Selection}.
While large models (in our benchmark GPT-4.1 and Claude Sonnet 4) perform the best across architectures, our benchmark shows that on simpler tasks, it is possible to achieve strong performances with smaller models if the appropriate architecture is chosen. For example, while GPT-4.1-mini's scores on the time off use case dip as low as 4.6\%, its peak acceptable score of 67.1\% is comparable to Sonnet 4's peak score of 68.5\%.

\section{Conclusion}
We introduce \textbf{AgentArch}, a benchmark for evaluating LLMs on a comprehensive agentic architectures on enterprise use cases. We test 18 configurations (single agent vs multi-agent, ReAct vs function calling, integration of thinking tools, and memory management styles) and find that even state-of-the-art LLMs struggle to maintain reliable performance on complex enterprise workflows in any setup. We also observe model-specific architectural preferences across LLMs, suggesting that there is no universally optimal agentic architecture. Furthermore, even the best performing models do not demonstrate reliability over multiple trials showing a fundamental gap between the potential of agentic systems and their effectiveness in practice. We hope these findings provide actionable insights that enterprise developers can use to improve their model selection and architecture design of agentic systems.

\section{Limitations}
Our benchmark has several limitations. 

\textbf{Use Case Diversity and Size}. We evaluate only two enterprise use cases with 60 samples each, which may not generalize across industries or contexts. One use case represents a simple workflow and the other a complex task, but more varied cases would better reveal how task complexity affects architectural performance. 

\textbf{Model selection}. Due to cost constraints, our experiments are currently limited to six models total, with only one open-source and one reasoning model. Including more reasoning models would clarify whether o3-mini’s results generalize to reasoning models as a class.

\textbf{Multimodal Enterprise Integration}. Our current evaluation is limited to text-based tool inputs and outputs, while real enterprise environments frequently involve multimodal data processing, including document analysis, image recognition, and file processing workflows. Future work should expand the benchmark to include tools that accept and return images, PDFs, and other file formats commonly encountered in enterprise systems. 

\textbf{Conversational Capabilities}
Our benchmark intentionally focuses on autonomous workflows without user interaction as this is an important and understudied application of agentic, particularly in enterprise contexts. However, many enterprise applications require dynamic user communication and multi-turn conversations. Future evaluations should incorporate both static settings as well as conversational settings where agents must clarify requirements and adapt to changing user needs throughout task execution. 

\textbf{Multi-Agent ReAct}. Our benchmark shows poor performance on multi-agent ReAct across models, but does not fully illuminate the reason for this poor performance. Future work focusing specifically on Multi-Agent ReAct and testing ReAct for orchestration and tool calling independently in multi-agent systems would provide more insights on whether ReAct can be effectively utilized in multi-agent systems.

\textbf{Sampling Parameters}. All experiments use temperature=0 for reproducibility, which may not reflect production deployments where some stochasticity could be beneficial. The interaction between sampling parameters and architectural choices remains unexplored.

\textbf{Success Metrics.} Our "Acceptable Score" requires perfect alignment on all three dimensions (tools, arguments, outcome). This strict evaluation may not capture scenarios where partial success provides business value. Additionally, we do not measure efficiency metrics (latency, cost, token usage) which are critical for enterprise deployment decisions.

\section{Acknowledgments}
We are grateful to Sai Rajeswar Mudumba for his careful reading of the manuscript and helpful suggestions and guidance that improved the final version of this work.

\bibliography{custom}

\appendix

\section{Use Cases and Set Up}
\label{sec:usecases}
When evaluating single agent modes, we simply concatenate all the agent instructions for each defined agent and put them within tags to indicate which step the instructions refer to. In this way, agents in the single agent setup have access to all the same information as in multi-agent setups. 

Within each use case, we represent a wide array of potential edge cases. For example, in the Requesting Time Off use case we have user requests that should be rejected and some that should be accepted. Within the rejected category, we include requests that should be rejected for a wide variety of reasons such as invalid leave types, conflicting leaves, and most significantly, lack of leave balance. For lack of balance, we make sure to include complex cases where models must correctly count the number of days across multiple months or take into account leap years and the differing number of days in different months. 

For each sample in a use case, the ground truth values for expected tool inputs, expected outcome, and expected tool order were human annotated and verified. We construct the use cases in such a way that a limited set of deterministic ground truths options can be determined for every data sample.

\section{Thinking Tools Examples}
Below are examples of how thinking tools can be used by the model to complete tasks in agentic frameworks. Thinking tools effectively give the model space to generate extra tokens in the format of calling a tool. There is no actual tool that gets executed when these tools are selected. Whatever the model passes as the tool argument is returned as the result and added to memory so the model can use the information later on in the task.

\begin{enumerate}
    \item \textbf{Synthesizing information}

    \textbf{tool:} \texttt{synthesize\_collected\_information} \\
    \textbf{args:}
    \begin{quote}
    \texttt{synthesized\_information (str):} \\
    \texttt{"I need to determine whether or not the employee Sarah Smith is eligible for leave. I see her leave balance is 18 days but she has already requested 4 days which leaves 14 days. Her current request is for 7 days and I see no leave conflicts and the type of leave is valid, which means she should be eligible for this current request."}
    \end{quote}
    \textbf{result (same as input):}
    \begin{quote}
    \texttt{"I need to determine whether or not the employee Sarah Smith is eligible for leave. I see her leave balance is 18 days but she has already requested 4 days which leaves 14 days. Her current request is for 7 days and I see no leave conflicts and the type of leave is valid, which means she should be eligible for this current request."}
    \end{quote}

    \item \textbf{Math}

    \textbf{tool:} \texttt{math} \\
    \textbf{args:}
    \begin{quote}
    \texttt{input (str):} \\
    \texttt{"Sarah Smith requested leave from 'all of February 2024 to the second of March'. I need to figure out how many days that is. Since February has 29 days in 2024 due to it being a leap year, that should be 29 days + 2 days in March, so 31 days."}
    \end{quote}
    \textbf{result  (same as input)::}
    \begin{quote}
    \texttt{"Sarah Smith requested leave from 'all of February 2024 to the second of March'. I need to figure out how many days that is. Since February has 29 days in 2024 due to it being a leap year, that should be 29 days + 2 days in March, so 31 days."}
    \end{quote}
\end{enumerate}

\section{AI Usage}
We use AI in this paper for generating the code required for running the benchmark, for proof reading and paraphrasing, and for creating the graphics and figure formatting.

\section{Prompts}
Prompts were designed to be simple and representative examples of both traditional function calling-style prompts and ReAct prompts. For function calling prompts, the tools and directions are placed in the system prompt while the user prompt includes the main objective and conversation history thus far. We also include a note to always call a tool (since communication with the user is not supported), always use real input values (since some models are prone to using placeholders instead), and to select the finish tool when finished since this is critical for control flow as well as getting a self reported summary of each agents actions which is passed to other agents in the summarized memory management style. We attempt to convey the same information in the ReAct style prompt however as is typical with ReAct prompting, all information is in the system prompt and a JSON response is required. The ReAct prompts include a sample JSON output format that asks the model for "thought" (reasoning) and an "action" and "observation.".

\section{Additional Metrics}
We include supplemental metrics here all shown with pass@1: Correct Tools Strict Rate, Correct Tools Lenient Rate, Tool Repetition Rate (higher scores indicate more repetition), Missing Required Tool Rate (higher scores indicate more missing tools). We also show Pass\^{}K for the Acceptable Rate.

\begin{table*}[t]
\centering
\resizebox{\linewidth}{!}{%
\begin{tabular}{llllccccccccccccc}
\toprule
Orchestration & Agent & Memory & Thinking Tools & & \multicolumn{2}{c}{GPT-4.1} & \multicolumn{2}{c}{GPT-4o} & \multicolumn{2}{c}{GPT-4.1-mini} & \multicolumn{2}{c}{o3-mini} & \multicolumn{2}{c}{LLaMA 3.3 70B} & \multicolumn{2}{c}{Sonnet 4} \\
\cmidrule(lr){6-7}
\cmidrule(lr){8-9}
\cmidrule(lr){10-11}
\cmidrule(lr){12-13}
\cmidrule(lr){14-15}
\cmidrule(lr){16-17}
& & &  & & TO & CR & TO & CR & TO & CR & TO & CR & TO & CR & TO & CR \\
\midrule
\multirow{6}{*}{Orch-Open} & \multirow{2}{*}{React} & Complete & Enabled &  & 8.1 & 19.5 & 14.4 & 1.2 & 13.3 & 2.5 & 5.8 & 0.0 & 4.2 & 0.0 & 14.0 & 0.0 \\
 &  & Complete & Disabled &  & 5.2 & 15.8 & 11.5 & 1.2 & 10.6 & 3.0 & 6.2 & 0.0 & 2.3 & 0.0 & 21.2 & 0.5 \\
\cmidrule(lr){3-17}
 & \multirow{4}{*}{Function Calling} & Complete & Enabled &  & 52.3 & 17.5 & 27.9 & 18.0 & 14.2 & 16.2 & 2.3 & 18.2 & 0.0 & 0.0 & 58.8 & 16.0 \\
 &  & Complete & Disabled &  & 49.6 & \textbf{21.5} & 17.1 & \textbf{19.8} & 12.1 & 16.8 & 2.1 & 19.8 & 0.0 & 0.0 & 49.8 & 17.2 \\
 &  & Summarized & Enabled &  & 49.0 & 17.5 & 24.2 & 17.0 & 15.4 & 19.0 & 2.9 & 16.8 & 0.0 & 0.0 & 61.3 & 16.8 \\
 &  & Summarized & Disabled &  & 49.6 & 17.5 & 24.0 & 16.8 & 6.2 & \textbf{21.5} & 2.7 & 17.0 & 0.0 & 0.0 & 57.1 & 16.5 \\
\midrule
\multirow{6}{*}{Orch-Isolated} & \multirow{2}{*}{React} & Complete & Enabled &  & 10.2 & 16.0 & 8.3 & 0.2 & 12.1 & 3.2 & 7.3 & 0.0 & \textbf{6.0} & 0.0 & 12.1 & 0.0 \\
 &  & Complete & Disabled &  & 9.8 & 15.2 & 9.8 & 0.0 & 10.4 & 1.2 & 5.0 & 0.0 & 3.3 & 0.0 & 18.8 & 0.0 \\
\cmidrule(lr){3-17}
 & \multirow{4}{*}{Function Calling} & Complete & Enabled &  & 50.0 & 17.5 & 47.1 & 13.8 & 6.0 & 0.8 & 2.5 & \textbf{20.2} & 0.0 & 0.0 & 61.0 & 15.0 \\
 &  & Complete & Disabled &  & 50.2 & 17.0 & \textbf{50.8} & 12.2 & 6.5 & 1.5 & 1.5 & 18.5 & 0.0 & 0.0 & 55.0 & 16.2 \\
 &  & Summarized & Enabled &  & 44.4 & 16.5 & 39.0 & 12.8 & \textbf{16.9} & 2.0 & 0.8 & 18.5 & 0.0 & 0.0 & \textbf{62.3} & 16.8 \\
 &  & Summarized & Disabled &  & 44.4 & 16.0 & 43.8 & 14.2 & 10.8 & 1.5 & 2.3 & 19.5 & 0.0 & 0.0 & 57.9 & 15.0 \\
\midrule
\multirow{6}{*}{Single Agent} & \multirow{2}{*}{React} & Complete & Enabled &  & 30.0 & 5.2 & 20.8 & 1.8 & 0.2 & 7.2 & 8.8 & 1.5 & 0.8 & 0.0 & 58.8 & 17.2 \\
 &  & Complete & Disabled &  & 39.4 & 9.8 & 13.1 & 1.0 & 0.0 & 5.2 & 4.0 & 2.2 & 0.8 & 0.0 & 47.9 & 31.0 \\
\cmidrule(lr){3-17}
 & \multirow{4}{*}{Function Calling} & Complete & Enabled &  & 68.5 & 12.2 & 31.9 & 4.2 & 0.0 & 0.2 & 38.5 & 1.2 & 0.0 & 0.0 & 3.3 & \textbf{37.2} \\
 &  & Complete & Disabled &  & 52.3 & 18.5 & 26.7 & 4.2 & 0.2 & 0.2 & 37.7 & 2.5 & 0.0 & 0.0 & 0.0 & 36.0 \\
 &  & Summarized & Enabled &  & \textbf{71.7} & 13.2 & 30.2 & 3.2 & 0.0 & 0.8 & 36.5 & 1.8 & 0.0 & 0.0 & 3.1 & 35.8 \\
 &  & Summarized & Disabled &  & 49.0 & 17.2 & 23.5 & 3.2 & 0.0 & 0.2 & \textbf{39.2} & 1.2 & 0.0 & 0.0 & 0.0 & 36.2 \\
\bottomrule
\end{tabular}%
}
\caption{\textbf{Pass@1 Correct Tools Strict Rate}. Bolded scores indicate highest score per model on use case}
\label{tab:results}
\end{table*}

\begin{table*}[t]
\centering
\resizebox{\linewidth}{!}{%
\begin{tabular}{llllccccccccccccc}
\toprule
Orchestration & Agent & Memory & Thinking Tools & & \multicolumn{2}{c}{GPT-4.1} & \multicolumn{2}{c}{GPT-4o} & \multicolumn{2}{c}{GPT-4.1-mini} & \multicolumn{2}{c}{o3-mini} & \multicolumn{2}{c}{LLaMA 3.3 70B} & \multicolumn{2}{c}{Sonnet 4} \\
\cmidrule(lr){6-7}
\cmidrule(lr){8-9}
\cmidrule(lr){10-11}
\cmidrule(lr){12-13}
\cmidrule(lr){14-15}
\cmidrule(lr){16-17}
& & &  & & TO & CR & TO & CR & TO & CR & TO & CR & TO & CR & TO & CR \\
\midrule
\multirow{6}{*}{Orch-Open} & \multirow{2}{*}{React} & Complete & Enabled &  & 37.9 & 21.8 & 26.5 & 7.0 & 32.3 & 19.5 & 8.8 & 0.2 & 16.2 & 4.5 & 28.1 & 30.8 \\
 &  & Complete & Disabled &  & 32.1 & 19.0 & 23.1 & 10.2 & 24.4 & 26.0 & 9.2 & 0.0 & 12.7 & 6.6 & 29.0 & 43.2 \\
\cmidrule(lr){3-17}
 & \multirow{4}{*}{Function Calling} & Complete & Enabled &  & 58.8 & 25.8 & 51.2 & 24.2 & 69.2 & 43.8 & 4.6 & 20.5 & 13.3 & 0.0 & 68.3 & 44.0 \\
 &  & Complete & Disabled &  & 54.8 & 28.5 & 46.5 & \textbf{26.8} & 56.9 & \textbf{51.7} & 5.4 & 21.8 & 14.2 & 0.0 & 57.7 & 46.0 \\
 &  & Summarized & Enabled &  & 55.6 & 23.2 & \textbf{55.4} & 25.5 & 68.3 & 42.2 & 5.2 & 18.5 & 0.0 & 0.0 & 69.8 & 42.0 \\
 &  & Summarized & Disabled &  & 55.2 & 25.2 & 51.2 & 24.2 & 54.8 & 46.8 & 5.2 & 19.2 & 0.0 & 0.0 & 57.9 & 43.2 \\
\midrule
\multirow{6}{*}{Orch-Isolated} & \multirow{2}{*}{React} & Complete & Enabled &  & 29.2 & 17.8 & 25.0 & 1.0 & 27.1 & 8.8 & 9.4 & 0.5 & \textbf{18.1} & 2.0 & 28.5 & 3.2 \\
 &  & Complete & Disabled &  & 26.2 & 17.0 & 24.2 & 1.8 & 26.9 & 7.5 & 8.3 & 0.2 & 13.5 & 0.8 & 28.1 & 2.2 \\
\cmidrule(lr){3-17}
 & \multirow{4}{*}{Function Calling} & Complete & Enabled &  & 58.8 & 17.8 & 53.5 & 14.0 & \textbf{70.4} & 0.8 & 6.0 & \textbf{22.2} & 3.1 & 0.0 & \textbf{70.2} & 42.8 \\
 &  & Complete & Disabled &  & 55.0 & 17.0 & 52.3 & 13.0 & 58.5 & 1.8 & 4.0 & 21.0 & 0.6 & 0.0 & 57.5 & 16.2 \\
 &  & Summarized & Enabled &  & 55.8 & 16.5 & 50.6 & 14.0 & 69.0 & 2.0 & 2.9 & 20.5 & 0.0 & 0.0 & 66.9 & 16.8 \\
 &  & Summarized & Disabled &  & 53.8 & 16.0 & 49.0 & 15.2 & 60.0 & 1.5 & 5.0 & 21.2 & 0.0 & 0.0 & 57.9 & 15.0 \\
\midrule
\multirow{6}{*}{Single Agent} & \multirow{2}{*}{React} & Complete & Enabled &  & 36.5 & 8.0 & 25.4 & 2.5 & 57.9 & 19.2 & 9.2 & 1.8 & 11.2 & 4.8 & 63.7 & 30.0 \\
 &  & Complete & Disabled &  & 46.0 & 16.0 & 19.8 & 3.5 & 59.4 & 13.0 & 4.2 & 2.5 & 12.7 & \textbf{7.9} & 54.2 & 35.8 \\
\cmidrule(lr){3-17}
 & \multirow{4}{*}{Function Calling} & Complete & Enabled &  & 69.0 & 29.0 & 40.6 & 5.5 & 43.5 & 12.2 & \textbf{56.9} & 1.5 & 0.0 & 0.0 & 66.7 & 43.5 \\
 &  & Complete & Disabled &  & 55.2 & \textbf{36.5} & 29.4 & 7.2 & 37.5 & 12.8 & 56.0 & 3.0 & 5.0 & 0.0 & 31.7 & \textbf{50.7} \\
 &  & Summarized & Enabled &  & \textbf{71.9} & 29.5 & 40.2 & 6.5 & 41.9 & 13.0 & 55.0 & 2.8 & 0.8 & 0.0 & 64.4 & 42.0 \\
 &  & Summarized & Disabled &  & 53.3 & 34.5 & 30.8 & 6.0 & 41.5 & 16.0 & 56.0 & 1.2 & 0.0 & 0.0 & 31.7 & 49.0 \\
\bottomrule
\end{tabular}%
}
\caption{\textbf{Pass@1 Correct Tools Lenient Rate}. Bolded scores indicate highest score per model on use case}
\label{tab:results}
\end{table*}

\begin{table*}[t]
\centering
\resizebox{\linewidth}{!}{%
\begin{tabular}{llllccccccccccccc}
\toprule
Orchestration & Agent & Memory & Thinking Tools & & \multicolumn{2}{c}{GPT-4.1} & \multicolumn{2}{c}{GPT-4o} & \multicolumn{2}{c}{GPT-4.1-mini} & \multicolumn{2}{c}{o3-mini} & \multicolumn{2}{c}{LLaMA 3.3 70B} & \multicolumn{2}{c}{Sonnet 4} \\
\cmidrule(lr){6-7}
\cmidrule(lr){8-9}
\cmidrule(lr){10-11}
\cmidrule(lr){12-13}
\cmidrule(lr){14-15}
\cmidrule(lr){16-17}
& & &  & & TO & CR & TO & CR & TO & CR & TO & CR & TO & CR & TO & CR \\
\midrule
\multirow{6}{*}{Orch-Open} & \multirow{2}{*}{React} & Complete & Enabled &  & 0.0 & 0.0 & 0.0 & 0.0 & 0.0 & 0.0 & 0.0 & 0.0 & 0.0 & 0.0 & 0.0 & 0.0 \\
 &  & Complete & Disabled &  & 0.0 & 0.0 & 0.0 & 0.0 & 0.0 & 0.0 & 0.0 & 0.0 & 0.0 & 0.0 & 0.0 & 0.0 \\
\cmidrule(lr){3-17}
 & \multirow{4}{*}{Function Calling} & Complete & Enabled &  & 1.4 & 0.0 & 0.4 & 0.0 & 3.5 & \textbf{0.0} & 0.0 & 0.0 & 0.0 & 0.0 & 4.4 & 0.0 \\
 &  & Complete & Disabled &  & 0.8 & 0.0 & 0.2 & \textbf{0.0} & 1.0 & \textbf{0.0} & 0.0 & 0.0 & 0.0 & 0.0 & 1.2 & 0.0 \\
 &  & Summarized & Enabled &  & 0.5 & 0.0 & 0.3 & 0.0 & 1.5 & 0.0 & 0.0 & 0.0 & 0.0 & 0.0 & 1.6 & 0.0 \\
 &  & Summarized & Disabled &  & 0.6 & 0.0 & 0.2 & 0.0 & 0.4 & 0.0 & 0.0 & 0.0 & 0.0 & 0.0 & 0.1 & 0.0 \\
\midrule
\multirow{6}{*}{Orch-Isolated} & \multirow{2}{*}{React} & Complete & Enabled &  & 0.0 & 0.0 & 0.0 & 0.0 & 0.0 & 0.0 & 0.0 & 0.0 & 0.0 & 0.0 & 0.0 & 0.0 \\
 &  & Complete & Disabled &  & 0.0 & 0.0 & 0.0 & 0.0 & 0.0 & 0.0 & 0.0 & 0.0 & 0.0 & 0.0 & 0.0 & 0.0 \\
\cmidrule(lr){3-17}
 & \multirow{4}{*}{Function Calling} & Complete & Enabled &  & 1.4 & 0.0 & \textbf{0.7} & 0.0 & \textbf{4.1} & 0.0 & 0.0 & \textbf{0.0} & 0.0 & 0.0 & \textbf{4.9} & 0.0 \\
 &  & Complete & Disabled &  & 0.8 & 0.0 & 0.5 & 0.0 & 1.1 & 0.0 & 0.0 & 0.0 & 0.0 & 0.0 & 1.2 & 0.0 \\
 &  & Summarized & Enabled &  & 0.8 & 0.0 & 0.2 & 0.0 & 1.6 & 0.0 & 0.0 & 0.0 & 0.0 & 0.0 & 1.3 & 0.0 \\
 &  & Summarized & Disabled &  & 0.5 & 0.0 & 0.1 & 0.0 & 0.8 & 0.0 & 0.0 & 0.0 & 0.0 & 0.0 & 0.2 & 0.0 \\
\midrule
\multirow{6}{*}{Single Agent} & \multirow{2}{*}{React} & Complete & Enabled &  & 0.0 & 0.0 & 0.0 & 0.0 & 0.5 & 0.0 & 0.0 & 0.0 & 0.0 & 0.0 & 2.7 & 0.0 \\
 &  & Complete & Disabled &  & 0.0 & 0.0 & 0.0 & 0.0 & 0.1 & 0.0 & 0.0 & 0.0 & \textbf{0.0} & 0.0 & 0.6 & 0.0 \\
\cmidrule(lr){3-17}
 & \multirow{4}{*}{Function Calling} & Complete & Enabled &  & 4.6 & \textbf{0.0} & 0.1 & 0.0 & 0.0 & 0.0 & \textbf{1.1} & 0.0 & 0.0 & 0.0 & 3.9 & 0.0 \\
 &  & Complete & Disabled &  & 0.4 & 0.0 & 0.0 & 0.0 & 0.0 & 0.0 & 0.9 & 0.0 & 0.0 & 0.0 & 0.0 & 0.0 \\
 &  & Summarized & Enabled &  & \textbf{6.3} & 0.0 & 0.0 & 0.0 & 0.0 & 0.0 & 0.8 & 0.0 & 0.0 & 0.0 & 2.9 & 0.0 \\
 &  & Summarized & Disabled &  & 0.3 & 0.0 & 0.0 & 0.0 & 0.0 & 0.0 & 0.9 & 0.0 & 0.0 & 0.0 & 0.0 & \textbf{0.0} \\
\bottomrule
\end{tabular}%
}
\caption{\textbf{Pass\^{} K Acceptable Rate}. Bolded scores indicate highest score per model on use case}
\label{tab:results}
\end{table*}

\begin{table*}[t]
\centering
\resizebox{\linewidth}{!}{%
\begin{tabular}{llllccccccccccccc}
\toprule
Orchestration & Agent & Memory & Thinking Tools & & \multicolumn{2}{c}{GPT-4.1} & \multicolumn{2}{c}{GPT-4o} & \multicolumn{2}{c}{GPT-4.1-mini} & \multicolumn{2}{c}{o3-mini} & \multicolumn{2}{c}{LLaMA 3.3 70B} & \multicolumn{2}{c}{Sonnet 4} \\
\cmidrule(lr){6-7}
\cmidrule(lr){8-9}
\cmidrule(lr){10-11}
\cmidrule(lr){12-13}
\cmidrule(lr){14-15}
\cmidrule(lr){16-17}
& & &  & & TO & CR & TO & CR & TO & CR & TO & CR & TO & CR & TO & CR \\
\midrule
\multirow{6}{*}{Orch-Open} & \multirow{2}{*}{React} & Complete & Enabled &  & 7.5 & 28.2 & 57.7 & 36.2 & 55.2 & 93.5 & \textbf{64.2} & 1.5 & 90.0 & 84.2 & 0.6 & \textbf{56.8} \\
 &  & Complete & Disabled &  & 9.4 & \textbf{35.2} & \textbf{58.1} & \textbf{41.8} & 53.8 & \textbf{94.0} & \textbf{64.2} & \textbf{3.2} & 94.6 & 96.8 & 0.0 & 40.2 \\
\cmidrule(lr){3-17}
 & \multirow{4}{*}{Function Calling} & Complete & Enabled &  & 9.4 & 0.0 & 9.4 & 5.0 & 9.6 & 17.5 & 0.0 & 0.0 & \textbf{100.0} & \textbf{100.0} & 0.0& 0.0 \\
 &  & Complete & Disabled &  & 11.0 & 0.0 & 10.6 & 3.8 & 15.6 & 16.8 & 0.0 & 0.0 & \textbf{100.0} & \textbf{100.0} & 0.0 & 0.0 \\
 &  & Summarized & Enabled &  & 12.5 & 0.0 & 3.8 & 2.8 & 24.0 & 19.5 & 0.0 & 0.0 & \textbf{100.0} & \textbf{100.0} & 0.0 & 0.0 \\
 &  & Summarized & Disabled &  & 13.1 & 0.0 & 4.2 & 1.8 & 31.0 & 18.8 & 0.0 & 0.0 & \textbf{100.0} & \textbf{100.0} & 0.0 & 0.2 \\
\midrule
\multirow{6}{*}{Orch-Isolated} & \multirow{2}{*}{React} & Complete & Enabled &  & 10.6 & 17.2 & 54.0 & 20.0 & 54.6 & 90.0 & 62.1 & 1.0 & 91.5 & 92.4 & \textbf{1.5} & 26.0 \\
 &  & Complete & Disabled &  & 10.2 & 22.8 & 46.2 & 28.7 & \textbf{61.3} & 81.8 & 62.7 & 2.0 & 95.6 & 91.2 & 0.4 & 24.2 \\
\cmidrule(lr){3-17}
 & \multirow{4}{*}{Function Calling} & Complete & Enabled &  & 6.9 & 0.2 & 1.7 & 7.8 & 9.8 & 18.2 & 0.0 & 0.0 & \textbf{100.0} & \textbf{100.0} & 0.0 & 0.0 \\
 &  & Complete & Disabled &  & 12.3 & 0.0 & 1.5 & 2.8 & 21.2 & 13.2 & 0.0 & 0.0 & \textbf{100.0} & \textbf{100.0} & 0.0 & 0.0 \\
 &  & Summarized & Enabled &  & 9.0 & 0.2 & 4.4 & 5.5 & 24.8 & 17.0 & 0.0 & 0.0 & \textbf{100.0} & \textbf{100.0} & 0.0 & 0.0 \\
 &  & Summarized & Disabled &  & \textbf{16.0} & 0.0 & 2.3 & 6.5 & 32.5 & 13.0 & 0.0 & 0.0 & \textbf{100.0} & \textbf{100.0} & 0.0 & 0.0 \\
\midrule
\multirow{6}{*}{Single Agent} & \multirow{2}{*}{React} & Complete & Enabled &  & 1.0 & 22.8 & 4.4 & 3.2 & 20.4 & 46.5 & 0.0 & 0.0 & 0.0 & 24.2 & 0.0 & 8.8 \\
 &  & Complete & Disabled &  & 1.9 & 21.2 & 2.9 & 5.5 & 9.4 & 29.8 & 0.0 & 0.0 & 0.4 & 12.7 & 0.0 & 8.0 \\
\cmidrule(lr){3-17}
 & \multirow{4}{*}{Function Calling} & Complete & Enabled &  & 0.0 & 5.8 & 0.2 & 6.0 & 0.2 & 4.5 & 0.0 & 0.0 & 0.0 & 88.9 & 0.0 & 0.0 \\
 &  & Complete & Disabled &  & 0.4 & 2.8 & 0.0 & 5.0 & 1.5 & 6.5 & 0.0 & \textbf{100.0} & 98.8 & \textbf{100.0} & 0.0 & 0.0 \\
 &  & Summarized & Enabled &  & 0.0 & 4.2 & 0.6 & 5.2 & 0.6 & 3.5 & 0.0 & \textbf{100.0} & 99.2 & \textbf{100.0}& 0.0 & 0.0 \\
 &  & Summarized & Disabled &  & 0.4 & 3.2 & 0.0 & 4.8 & 1.7 & 5.0 & 0.0 & 0.0 & 0.0 & 98.8 & 0.0 & 0.0 \\
\bottomrule
\end{tabular}%
}
\caption{\textbf{Pass@1 Tool Repetition Rate ↓.} Bolded scores indicate highest score (most repetition) per model on use case}
\label{tab:results}
\end{table*}

\begin{table*}[t]
\centering
\resizebox{\linewidth}{!}{%
\begin{tabular}{llllccccccccccccc}
\toprule
Orchestration & Agent & Memory & Thinking Tools & & \multicolumn{2}{c}{GPT-4.1} & \multicolumn{2}{c}{GPT-4o} & \multicolumn{2}{c}{GPT-4.1-mini} & \multicolumn{2}{c}{o3-mini} & \multicolumn{2}{c}{LLaMA 3.3 70B} & \multicolumn{2}{c}{Sonnet 4} \\
\cmidrule(lr){6-7}
\cmidrule(lr){8-9}
\cmidrule(lr){10-11}
\cmidrule(lr){12-13}
\cmidrule(lr){14-15}
\cmidrule(lr){16-17}
& & &  & & TO & CR & TO & CR & TO & CR & TO & CR & TO & CR & TO & CR \\
\midrule
\multirow{6}{*}{Orch-Open} & \multirow{2}{*}{React} & Complete & Enabled &  & 36.0 & 65.5 & 32.3 & 62.0 & 32.3 & 58.8 & 74.6 & 63.2 & 31.0 & 68.2 & 3.1 & 15.5 \\
 &  & Complete & Disabled &  & 43.5 & 65.8 & 34.6 & 60.8 & 36.9 & 50.5 & 70.4 & 62.3 & 41.9 & 85.4 & 3.3 & 5.8 \\
\cmidrule(lr){3-17}
 & \multirow{4}{*}{Function Calling} & Complete & Enabled &  & 13.8 & 58.5 & 5.0 & 56.2 & 6.2 & 27.3 & 94.6 & 74.2 & 85.8 & \textbf{100.0} & 0.6 & 40.0 \\
 &  & Complete & Disabled &  & 9.0 & 56.5 & 2.7 & 54.2 & 5.0 & 16.2 & 94.0 & 73.8 & 83.5 & \textbf{100.0} & 3.8 & 40.0 \\
 &  & Summarized & Enabled &  & 7.1 & 61.3 & 4.2 & 54.8 & 5.4 & 27.8 & 94.2 & 74.2 & \textbf{100.0} & \textbf{100.0}& 2.1 & 40.0 \\
 &  & Summarized & Disabled &  & 6.5 & 60.0 & 3.3 & 55.2 & 6.9 & 18.2 & 94.0 & 73.5 & \textbf{100.0} & \textbf{100.0} & 1.9 & 40.0 \\
\midrule
\multirow{6}{*}{Orch-Isolated} & \multirow{2}{*}{React} & Complete & Enabled &  & 55.2 & \textbf{70.0} & 40.0 & 72.8 & \textbf{43.8} & 70.2 & 76.2 & 63.7 & 28.5 & 69.2 & 2.3 & 66.8 \\
 &  & Complete & Disabled &  & \textbf{57.5} & 69.8 & 42.5 & 72.8 & 39.0 & \textbf{70.5} & 73.3 & 62.3 & 42.1 & 77.8 & 2.7 & 68.2 \\
\cmidrule(lr){3-17}
 & \multirow{4}{*}{Function Calling} & Complete & Enabled &  & 13.5 & 69.8 & 10.8 & 69.8 & 5.0 & 70.0 & 92.5 & 73.0 & 96.9 & 96.4 & 0.6 & 40.2 \\
 &  & Complete & Disabled &  & 16.7 & \textbf{70.0} & 1.9 & 69.2 & 4.4 & 70.0 & 95.0 & 72.8 & 99.4 & 98.2 & 2.5 & \textbf{70.0} \\
 &  & Summarized & Enabled &  & 19.4 & \textbf{70.0} & 11.2 & 68.8 & 4.0 & 70.0 & \textbf{95.6} & 73.0 & \textbf{100.0} & 98.8 & 3.1 & \textbf{70.0} \\
 &  & Summarized & Disabled &  & 17.3 & \textbf{70.0} & 5.0 & 69.0 & 2.9 & 70.0 & 94.2 & 74.0 & \textbf{100.0} & \textbf{100.0} & 3.5 & \textbf{70.0} \\
\midrule
\multirow{6}{*}{Single Agent} & \multirow{2}{*}{React} & Complete & Enabled &  & 34.0 & 62.0 & 40.8 & 74.0 & 23.5 & 57.0 & 44.6 & 83.5 & 47.1 & 79.2 & 10.6 & 33.8 \\
 &  & Complete & Disabled &  & 32.9 & 55.2 & \textbf{46.7} & \textbf{75.8} & 28.3 & 62.0 & 34.8 & 77.8 & 51.9 & 82.4 & \textbf{16.7} & 28.5 \\
\cmidrule(lr){3-17}
 & \multirow{4}{*}{Function Calling} & Complete & Enabled &  & 3.5 & 41.0 & 18.8 & 65.0 & 16.9 & 57.8 & 24.8 & \textbf{94.8} & \textbf{100.0} & \textbf{100.0} & 10.2 & 22.5 \\
 &  & Complete & Disabled &  & 16.5 & 33.5 & 24.6 & 62.7 & 18.8 & 57.8 & 26.2 & 92.5 & 91.9 & \textbf{100.0} & 11.5 & 19.5 \\
 &  & Summarized & Enabled &  & 3.1 & 40.5 & 18.3 & 65.0 & 16.5 & 56.5 & 25.4 & 93.5 & 98.3 & \textbf{100.0} & 12.5 & 24.2 \\
 &  & Summarized & Disabled &  & 17.3 & 35.5 & 18.3 & 64.8 & 15.4 & 53.5 & 24.8 & 94.2 & \textbf{100.0} & \textbf{100.0} & 11.7 & 20.2 \\
\bottomrule
\end{tabular}%
}
\caption{\textbf{Pass@1 Missing Required Tool Rate ↓}. Bolded scores indicate highest score (most missing tools) per model on use case}
\label{tab:results}
\end{table*}

\begin{samepage}
\begin{figure*}[!htb]
\fbox{\begin{minipage}{\dimexpr\linewidth-2\fboxsep-2\fboxrule\relax}
{\small\ttfamily
\textbf{system}:\\\\
You are an agent. Your goal is to accomplish your main objective using the directions provided. You must follow the directions you are given. If the directions instruct you to complete steps in a specific order, you must follow that order.\\\\
    
Directions:
\{agent\_instructions\}\\\\
\textbf{user}:\\\\
Main Objective:
\{main\_objective\}\\\\
    
Conversation History:
\{memory\}\\\\
    
Always call a tool. You must always pass real values as tool inputs. You may never use placeholders.
Before calling a tool, remember to consider the conversation history to understand what has already been accomplished and what needs to be done next.
When you have finished your task, you MUST select the finish tool. Otherwise your response will be ignored.
}
\end{minipage}}
\caption{Function Calling Agent Prompt}
\end{figure*}
\end{samepage}

\begin{samepage}
\begin{figure*}[!htb]
\fbox{\begin{minipage}{\dimexpr\linewidth-2\fboxsep-2\fboxrule\relax}
{\small\ttfamily
\textbf{user}:
     You are an agent. Your goal is to accomplish your main objective using the directions provided. You must follow the directions you are given. If the directions instruct you to complete steps in a specific order, you must follow that order.\\\\
    
    Directions:\\
    \{agent\_instructions\}\\
    
    Main Objective:\\
    \{main\_objective\}\\
    
    Conversation History:\\
    \{memory\}\\
    
    To accomplish your main objective, you must select an action from the following options:\\
    \{tools\}\\
    
    Always select an action using the format below. You must always pass real values as arguments. You may never use placeholders.
    Before choosing an action, remember to consider the conversation history to understand what has already been accomplished and what needs to be done next.
    When you have finished your task, you MUST select the finish action. Otherwise your response will be ignored.\\\\
    
    Output your response in the following JSON format:\\
    {[}\\
    \{\\
    "thought": "<reason step by step about the current situation and what to do next>",\\
    "action": \{ \\
    "name": "<tool\_name\_1>", \\
    "arguments": \{ \\
    "<argument\_name>": <argument\_value> \\
           \} \\
          \} \\
        "observation": "<expected result of the tool execution>" \\
      \}... \\
      \{ \\
          "thought": "<reason step by step about the current situation and what to do next>",\\
          "action": \{\\
              "name": "<tool\_name\_m>",\\
              "arguments": \{\\
                  "<argument\_name>": <argument\_value>\\
              \}\\
            \}\\
          "observation": "<expected result of the tool execution>"\\
      \}\\
    {]}\\
}
\end{minipage}}
\caption{ReAct Agent Prompt}
\end{figure*}
\end{samepage}

\begin{samepage}
\begin{figure*}[!htb]
\fbox{\begin{minipage}{\dimexpr\linewidth-2\fboxsep-2\fboxrule\relax}
{\small\ttfamily
\textbf{user}: 
You are the orchestrator of a group of agents. Your job is to select the next agent(s) best suited for accomplishing the request given the conversation history thus far. You must follow the directions you are given. If the directions instruct you to complete steps in a specific order, you must follow that order.\\
The conversation history below shows you what agents have already done to accomplish the task.\\

Directions:\\
\{usecase\_directions\}\\

Request:\\
\{request\}\\

Conversation History:\\
\{memory\}\\

To accomplish your request, you must select from the following agents:\\
\{agents\}\\

Always select an agent using the format below. You must always pass real values as arguments. You may never use placeholders. 
Before selecting the next agent(s), consider your request, directions, and the conversation history thus far to understand what to do next.\\

Output your response in the following JSON format:\\
{[}\\
\{\\
"thought": "<reason step by step about the current situation and what to do next>",\\
"action": \{\\
"name": "<agent\_name\_1>",\\
"arguments": \{\\
"main\_objective": <main objective you want the agent to accomplish>\\
\}\\
\}\\
"observation": "<expected result of the agent execution>"\\
\}...\\
\{\\
"thought": "<reason step by step about the current situation and what to do next>",\\
"action": \{\\
"name": "<agent\_name\_n>",\\
"arguments": \{\\
"main\_objective": <main objective you want the agent to accomplish>\\
\}\\
\}\\
"observation": "<expected result of the agent execution>"\\
\}\\
{]}

If there are no relevant agents, respond with "Unfortunately no agents are able to help with this request." Otherwise you MUST respond with at least one agent in the format above.
}.
\end{minipage}}
\caption{Orchestrator ReAct Prompt}
\end{figure*}
\end{samepage}

\begin{samepage}
\begin{figure*}[!htb]
\fbox{\begin{minipage}{\dimexpr\linewidth-2\fboxsep-2\fboxrule\relax}
{\small\ttfamily
\textbf{system}:\\
    You are the orchestrator of a group of agents. Your job is to select the next agent(s) best suited for accomplishing the request. You must follow the directions you are given. If the directions instruct you to complete steps in a specific order, you must follow that order.
    When you select an agent, you must give it a main objective that describes what the agent should accomplish. \\\\

    You also have access to the conversation history below which shows you what agents have already done to accomplish the task. \\\\
    
    Directions:
    \{usecase\_directions\}\\\\
\textbf{user}:\\
Request:
\{user\_utterance\}\\\\
    
    Conversation History:
    \{memory\}\\\\

    Always select an agent. You must always pass real values as arguments. You may never use placeholders. 
    Before selecting the next agent(s), consider your request, directions, and the conversation history thus far to understand what to do next. 
    If there are no relevant agents, respond with "Unfortunately no agents are able to help with this request." Otherwise you MUST respond with at least one agent
}.
\end{minipage}}
\caption{Orchestrator ReAct Prompt}
\end{figure*}
\end{samepage}
\end{document}